\documentclass[11pt]{article}
\usepackage[a4paper,top=2cm,bottom=2cm,left=3cm,right=3cm,marginparwidth=1.75cm]{geometry}
\usepackage[utf8]{inputenc}
\usepackage[french,english]{babel}
\usepackage[T1]{fontenc}
\usepackage{booktabs}
\usepackage{graphicx}
\usepackage{amsmath}
\usepackage[dvipsnames]{xcolor}
\usepackage[colorlinks=true, allcolors=blue]{hyperref}
\usepackage{booktabs}
\usepackage[size=scriptsize,textwidth=25mm]{todonotes}
\usepackage{verbatim}
\usepackage[autolanguage,np]{numprint}
\usepackage{subcaption}
\usepackage{soul} 
\usepackage{eurosym}
\usepackage{dirtree}
\usepackage{authblk}

\usepackage[numbib,nottoc]{tocbibind} 

\def\corpusabbr{PARHAF}

\title{\corpusabbr, a human-authored corpus of clinical reports \\for fictitious patients in French}
\author[1]{Xavier Tannier}
\affil[1]{Sorbonne Université, Université Sorbonne Paris Nord, Inserm, Limics, F-75006 Paris, France}
\author[2,3]{Salam Abbara}
\affil[2]{Université Paris-Saclay, UVSQ, Assistance Publique-Hôpitaux de Paris, Raymond Poincaré University Hospital, Infectious Disease Department, Garches, France}
\affil[3]{Yonsei University College of Medicine, Gangnam Severance Hospital, Department of Laboratory Medicine, Seoul, South Korea}
\author[4]{Rémi Flicoteaux}
\affil[4]{Assistance Publique-Hôpitaux de Paris, Department of medical information, Paris, France}
\author[5]{Youness Khalil}
\affil[5]{Health Data Hub, 75015, Paris, France}
\author[6]{Aurélie Névéol}
\author[6]{Pierre Zweigenbaum}
\affil[6]{Université Paris-Saclay, CNRS, LISN, 91400, Orsay, France}
\author[5,7]{Emmanuel Bacry}
\affil[7]{Université Paris-Dauphine, PSL, CNRS, CEREMADE, 75016, Paris, France}

\date{}

\begin{document}
\maketitle

\newcommand{\AN}[1]{\todo[color=Turquoise,size=tiny]{[AN] #1}}
\newcommand{\ANrem}[1]{\todo[color=Turquoise,inline,caption={}]{[AN] #1}}
\newcommand{\ANi}[1]{\textcolor{teal}{#1}}
\newcommand{\XT}[1]{\todo[color=olive,size=tiny]{[XT] #1}}
\newcommand{\XTrem}[1]{\todo[color=olive,inline,caption={}]{[XT] #1}}
\newcommand{\XTi}[1]{\textcolor{olive}{#1}}
\newcommand{\PZ}[1]{\todo[color=Lavender, author=PZ,size=tiny]{#1}}
\newcommand{\PZrem}[1]{\todo[color=Lavender,inline,caption={}, author=PZ]{#1}}
\newcommand{\PZi}[1]{\textcolor{Mulberry}{#1}}
\newcommand{\SArem}[1]{\todo[color=Apricot, inline, author=SA]{#1}}
\newcommand{\EB}[1]{\todo[color=cyan,size=tiny]{[EB] #1}}
\newcommand{\EBrem}[1]{\todo[color=cyan,inline,caption={}]{[EB] #1}}
\newcommand{\EBi}[1]{\textcolor{blue}{#1}}
\newcommand{\FPrem}[1]{\todo[color=gray,inline,caption={}]{[FP] #1}}
\newcommand{\FPi}[1]{\textcolor{gray}{#1}}
\newcommand{\XFPrem}[1]{\todo[color=Apricot, inline, author=XFP]{#1}}

\def\patientcountall{\numprint{5009}}
\def\doccountall{\numprint{7394}}
\def\patientcounttrain{\numprint{4254}}
\def\doccounttrain{\numprint{6185}}
\def\patientcounttest{\numprint{755}}
\def\doccounttest{\numprint{1209}}

\small{Corresponding author: Xavier Tannier, \texttt{xavier.tannier@sorbonne-universite.fr}}

\begin{abstract}
The development of clinical natural language processing (NLP) systems is severely hampered by the sensitive nature of medical records, which restricts data sharing under stringent privacy regulations, particularly in France and the broader European Union. To address this gap, we introduce PARHAF, a large open-source corpus of clinical documents in French. PARHAF comprises expert-authored clinical reports describing realistic yet entirely fictitious patient cases, making it anonymous and freely shareable by design. The corpus was developed using a structured protocol that combined clinician expertise with epidemiological guidance from the French National Health Data System (SNDS), ensuring broad clinical coverage. A total of 104 medical residents across 18 specialties authored and peer-reviewed the reports following predefined clinical scenarios and document templates.

The corpus contains \doccountall\ clinical reports covering \patientcountall\ patient cases across a wide range of medical and surgical specialties. It includes a general-purpose component designed to approximate real-world hospitalization distributions, and four specialized subsets that support information-extraction use cases in oncology, infectious diseases, and diagnostic coding. Documents are released under a CC-BY open license, with a portion temporarily embargoed to enable future benchmarking under controlled conditions.

PARHAF provides a valuable resource for training and evaluating French clinical language models in a fully privacy-preserving setting, and establishes a replicable methodology for building shareable synthetic clinical corpora in other languages and health systems.

\end{abstract}

\newpage

\section{Background \& Summary}


\subsection{Context and Motivation}

Much of the information in electronic health records is conveyed by text such as clinical notes and discharge summaries (see, e.g.,~\cite{Laparra:YMI2021}).  Natural language processing aims to unlock that information and make it available for downstream tasks.  Publicly available clinical text corpora are a key asset to design, tune, and evaluate clinical natural language processing systems~\cite{Gao:JAMIA2022}.
Sharing clinical text is, however, difficult: the tension between individual data privacy and corpus distributability has been widely acknowledged as the central obstacle to making clinical corpora publicly available [3].

Some resources have been made available for U.S. English clinical NLP over the past few decades~\cite{Gao:JAMIA2022}, starting with the 2007 Computational Medicine Challenge~\cite{pestian-etal-2007-shared} and the i2b2 series of clinical NLP shared tasks~\cite{Uzuner:JAMIA2007}, many of which relied on the MIMIC database~\cite{Saeed:CCM2011,Johnson:SD2016}. However, access hurdles remain particularly salient in the French context. The European regulatory framework, among the most protective of health data, imposes severe restrictions on the circulation and secondary use of medical records.
This creates a marked scarcity of open and usable corpora of French medical reports~\cite{neveol2018clinical}. 

Beyond data access, models trained on clinical reports may themselves become sensitive, as they can memorize patient information during training, making the sharing of trained models legally and ethically challenging~\cite{berthelier2023toward}. Together, these factors create a fragmented ecosystem in which institutions and research teams operate in isolation, unable to effectively pool data or models. This combination of restricted data access, model sensitivity, lack of open resources, and resulting fragmentation severely limits the development and robust evaluation of NLP systems applied to French clinical text.

This creates the following challenge: 
\emph{Given this privacy bottleneck, how can a large, realistic, and fully privacy-preserving corpus of clinical reports be created to help clinical language processing research and development while being freely shareable?}

To address this challenge, NLP researchers have studied de-identification methods that remove personally identifying information from original clinical text \cite{Uzuner:JAMIA2007,Neamatullah:BMCMIDM2008,Grouin:MIE2009,Dernoncourt:JAMIA2016,Chevrier:JMIR2019} and used them to de-identify clinical datasets such as that in MIMIC \cite{Neamatullah:BMCMIDM2008}.  However, the resulting text is pseudonymized (directly identifying information has been removed) but not anonymized (there is no guarantee that reidentification is impossible%
).  This prevents it from being freely distributed.  In the United States, the MIMIC database can be shared under a stringent data-use agreement, but it remains unclear whether this protocol is compatible with E.U. regulation. 
For this reason, clinical document collections in French (e.g., the MERLOT corpus~\cite{campillos2018french} and other corpora extracted from French clinical data warehouses~\cite{jannot2017georges,tannier2024development}) were used in evaluation studies with explicit targeted ethical board approval but could not be shared due to privacy restrictions.

Hahn~\cite{HahnJAMIAO2025} notes that, faced with the non-shareability of real patient records, NLP researchers have developed a variety of proxies for clinical text.
One type of proxy is machine translation of English clinical datasets: Becker et al.~\cite{Becker:2016} translated into German the ShARe/CLEF eHealth 2013 training dataset based on MIMIC-II data \cite{Saeed:CCM2011}, Neves et al. ~\cite{neves2022findings} translated some clinical cases from English into French (with a focus on evaluating the performance of machine translation for measures and acronyms) and Frei et al.~\cite{Frei:JBI2023a} translated into German the 2018 n2c2 shared task dataset that reused data from MIMIC-III \cite{Johnson:SD2016}.  However, translated text requires thorough human review, and cultural and health system differences make the resulting text sensibly different from native clinical text.

Another proxy is synthetic clinical text.
GraSCCo \cite{Modersohn:GMDS2022} manually edited 63 deidentified German discharge summaries and case reports at multiple linguistic levels to make reidentification virtually impossible.
Recent efforts have also explored the use of autoregressive generative language models to produce synthetic clinical documents in English~\cite{iveNature2020}, French~\cite{hiebelEACL2023}, German~\cite{Frei:JBI2023b}, Swedish and Spanish~\cite{vakili-etal-ACL2025}. Nonetheless, the balance between privacy and utility of the resulting material needs further analysis~\cite{melamudClinicalNLP2019,estignardAIME2025}.   

Published case reports are a more distant proxy for clinical text, but their open-source status and existence in multiple languages have made them particularly attractive for clinical NLP.  Case reports have been collected, for instance, in the following corpora: CAS~\cite{grabar2018cas} (French), E3C~\cite{magnini2020e3c} (Italian, English, French, Spanish, and Basque), CANTEMIST~\cite{Miranda-Escalada:SEPLN2020} and DISTEMIST~\cite{Miranda-Escalada:CLEF2022} (Spanish), and FRASIMED~\cite{zaghir-etal-2024-frasimed} (French translations of CANTEMIST and DISTEMIST). The style of case reports, however, is quite different from that of electronic health records.

The closest proxy for true clinical texts is those written by health care professionals about fictitious patients, for instance, in medical textbooks or course material. The JSynCC corpus~\cite{Lohr:LREC2018} extracted 400 operative reports and 470 case reports from such textbooks.  The initial copyright on the textbooks, though, prevents the free distribution of the corpus.  The PARROT corpus~\cite{LEGUELLEC2026100066} contains 2,658 radiology reports about fictitious patients, including 475 in French, written on a volunteer basis by healthcare professionals from 21 countries. This endeavor was made possible through human networking, including leveraging professional radiological societies, which may be difficult to scale up to a diversity of medical specialties.

\subsection{Objectives and Contributions}

To overcome this limitation, the approach adopted in the present work was to ask healthcare professionals to write new clinical reports describing fictitious patients specifically for the creation of a shareable corpus, and to distribute these reports under an open license. Because the reports are created for this purpose and do not derive from real patient data, they are anonymous and shareable by design.
However, this approach raises important methodological questions: how can such reports be generated in a way that ensures both medical realism and statistical representativeness while preserving privacy?
To address this challenge, we designed a corpus creation protocol that leverages clinicians’ expertise while being guided by public health statistics, principles of corpus development \cite{AHDS:2004,Zweigenbaum:SHTI2001}, and a set of predefined clinical scenarios. The protocol was implemented using a large pool of French-speaking clinicians through a partnership with associations and unions of medical residents across multiple medical and surgical specialties, which recruited 104 residents as report authors. Guidelines were developed for selecting clinical cases, using data from the French National Health Data System (SNDS~\cite{moore2021national}) as reference scenarios for report creation, and the residents authored synthetic medical reports following these guidelines. The resulting open-source French-language corpus can now be used to train and evaluate language models on targeted medical use cases.

In this article, we introduce this open-source corpus of French clinical documents. \corpusabbr\  comprises \doccountall\ expert-authored clinical reports describing \patientcountall\ realistic yet fictitious patient cases. Each case is accompanied by structured documentation of the underlying clinical scenario, including the primary diagnosis, main procedure, care pathway, and discharge information when applicable. We further provide three specialized subsets specifically designed to support information extraction tasks in oncology and infectious diseases. This corpus offers a valuable resource for the development and evaluation of clinical NLP models, directly tackling the root cause of all the challenges outlined above : the inherently sensitive nature of clinical data.

We release \doccounttrain\ documents corresponding to \patientcounttrain\ fictitious patients under an open-source CC-BY license. The remaining portion of the corpus will be temporarily embargoed to enable future evaluations under controlled conditions, thereby limiting the risk of large language model contamination through prior exposure to the data.

\subsection{Intended uses of \corpusabbr}

This corpus is intended for research, development, and educational purposes in clinical natural language processing. It enables the sharing of clinical-style notes and annotations and supports community-wide pooling of efforts around a common, openly accessible resource. The corpus is suitable for benchmarking French medical language models, including large language models, and for conducting reproducible clinical NLP research under controlled and privacy-safe conditions.

The corpus also supports uses for medical teaching, such as training medical students and residents in structured clinical report writing, diagnostic reasoning, and clinical information synthesis. It can also serve as a resource for clinical case preparation and supports training in clinical natural language processing using realistic yet fictitious reports without exposure to sensitive patient data. The corpus further enables privacy-preserving data augmentation, either as a standalone resource or as a complement to restricted-access clinical datasets, provided its fictitious nature is explicitly acknowledged.

Finally, the representativeness of part of the corpus is geared towards three use cases of the PARTAGES project, allowing methodological comparisons across these specific use cases.

\subsection{Limitations and non-intended uses}

Although efforts were made to create a diverse corpus that includes a variety of document types and clinical specialties, the corpus does not cover all specialties and variations of French clinical text. 

This corpus is intended for research purposes only, specifically for training and evaluating natural language processing models on French clinical text. It is not a substitute for clinically validated data and must not be used to support regulatory approval, clinical certification, or deployment decisions in real healthcare settings.
It is not suitable for clinical use.
It cannot be used for clinical decision-making, diagnosis, prognosis, treatment, or patient care. Models trained or evaluated on this data are not clinically validated, and results obtained on this corpus cannot be presented as evidence of clinical performance or safety.

The corpus does not support generalization claims to real hospitals, regions, or clinical practices, nor does it allow epidemiological or population-level inference, as its distributions do not reflect real-world prevalence. It is also unsuitable for longitudinal studies or for assessing real-world clinical risk or safety, including rare adverse events or edge cases, and must not be used as a replacement for real clinical data in deployment settings. Finally, the corpus does not capture the operational constraints of real clinical environments (e.g., time pressure, workload, interruptions) and should not be used for stress-testing models under realistic clinical conditions.

\section{Methods}

\subsection{Challenges}

Building the \corpusabbr\ corpus required addressing two main challenges. The first was ensuring that recruited physicians and collected texts adequately represent the relevant dimensions of clinical language, while remaining within concrete implementation constraints (a limited author pool, a fixed corpus size, and the specific use cases targeted by the project). The second was encouraging healthcare professionals to write reports that closely resemble real clinical documents, while minimizing the risk of privacy leaks.





\subsection{Clinical Scenario Design}

For the above reason, we deemed essential to provide relatively precise guidelines to assist healthcare professionals in authoring clinical reports that closely resemble real-world documents while minimizing the risk of privacy breaches. These guidelines addressed both the content and the format of the documents. Given that the recruited physicians primarily worked in hospital settings, the resulting corpus of documents focused predominantly on hospital-based clinical situations.

\subsubsection{Content Development}

The selection of clinical scenarios was guided by our goal of guaranteeing the representativeness of the clinical situations actually observed in French hospitals (see Section~\ref{sec:core_distribution}) and by the constraint of ensuring physicians were familiar with the clinical situation in relation to their specialty of practice or training.
Both aspects were addressed
using hospitalization claims data available in the French National Health Data System (SNDS~\cite{moore2021national}).

Scenarios were constructed by sampling observed distributions of Diagnosis-Related Groups (DRGs), principal diagnoses (ICD-10), age, sex, type of management (e.g., ambulatory surgery), admission and discharge modes (e.g., emergency department admission). DRGs were used (in a less formal format) to describe the type of hospitalization (e.g., surgery, medicine) and to map clinical cases to physicians' qualifications (specialties). Secondary diagnoses were incorporated into the scenarios as a list of 10 randomly selected diagnoses from the pool of diagnoses frequently associated with the primary diagnosis-DRG pair. Patient name was randomly fixed. 

Based on these core elements, authors were encouraged to develop the clinical case details, enriching the content with relevant and realistic information, maintaining medical consistency with the baseline information, and ensuring depth and authenticity while adhering to principles of plausibility and ethics.

\subsubsection{Document Format}
For document format, we aligned 
official recommendations with physicians’ actual practices to develop specialized templates for each type of hospitalization:

\begin{itemize}
\item Medical hospitalization
\begin{itemize}
\item Hospital discharge summary
\end{itemize}
\item Surgical hospitalization
\begin{itemize}
\item Pre-operative consultation report (for scheduled admissions)
\item Operative report
\item Hospital discharge summary (for ambulatory surgery, a single document was requested combining both the operative report and the discharge summary)
\end{itemize}
\item Obstetrics (childbirth)
\begin{itemize}
\item Pre-delivery hospitalization report (for high-risk pregnancies) or emergency department visit report (for low-risk pregnancies)
\item Delivery room report
\item Postpartum hospitalization report (maternity ward)
\end{itemize}
\item Oncology
\begin{itemize}
\item Pathology report 
\end{itemize}
\end{itemize}

For discharge summaries, the template included: department name, reason for admission, medical history, surgical history, family history, allergies, lifestyle factors, treatment at admission, history of the present illness, clinical examination, complementary investigations, in-hospital course, discharge treatment, and conclusion. Similar minimal templates were developed for surgeries, obstetrics, and pathology reports. Authors were encouraged to follow these structures or to write in free text format, provided that all required information was included.

Finally, a structured summary section was completed at the end of each report, in which the authors specified the primary diagnosis, length of stay, and associated diagnoses mentioned in the report.

The use of generative artificial intelligence tools was discouraged because it could bias both the content and the stylistic features of the reports.


\subsection{Document Type and Distribution Strategy}

This corpus is structured in two complementary components, targeting a total of \numprint{5000} patients. The primary component (n = \numprint{3900}) includes patients across a wide range of medical specialties and is designed to maximize diversity and approximate representativeness, although the target size does not allow full coverage of the spectrum of possible clinical cases. The secondary component focuses on specific use cases (ICD-10 coding, oncology, and infectious diseases) and comprises patients selected outside the main distribution to support more targeted evaluation scenarios.

\subsubsection{Core distribution}
\label{sec:core_distribution}


To approximate real-world distributions of medical activity, we relied on diagnosis frequencies derived from SNDS~\cite{moore2021national}, which provides exhaustive, nationwide hospital claims data and served as a proxy for the underlying epidemiological and care distribution across medical conditions. 
For the year 2024, the national claims database consisted of approximately 18~million hospitalizations drawn from the SNDS. From the data, we defined clinical cases as the association of a DRG, sex, age group, and length-of-stay group. With these associations, we created a sampling database of \numprint{100000} different clinical cases.
These cases covered around \numprint{4000} distinct ICD-10 primary diagnoses. To ensure patient privacy and data confidentiality, the sampling strategy over this clinical cases distribution adheres to the principles of k-anonymity and l-diversity \cite{Sweeney2002kAnonymityAM}.


To preserve epidemiological realism while avoiding excessive over-representation of very frequent conditions, which would reduce clinical diversity in the corpus, we applied a square-root transformation to the empirical frequencies, yielding a preliminary sampling probability proportional to $\sqrt{f_i}$, where $f_i$ is the frequency of condition $i$ in the SNDS data. To further limit the dominance of the most common conditions, we capped this value at a maximum probability $p_{\mathrm{max}}$ (corresponding to a \numprint{0.1}\% sampling chance) and renormalized, giving the final sampling probability for each condition: 
$$p_i = \frac{\min(p_{\mathrm{max}},\, \sqrt{f_i})}{\sum_j \min(p_{\mathrm{max}},\, \sqrt{f_j})}$$

In practice, this theoretical distribution required iterative adjustment to account for operational constraints. Because hired authors had uneven expertise across medical specialties, document production could not be distributed uniformly, and not all specialties could be covered. The final allocation therefore used the square-root-with-cap model as a guiding principle, with reallocation based on actual case availability and author capacity, while preserving broad clinical coverage. 
Figures~\ref{fig:change_in_distribution} and~\ref{fig:count_per_author} in the Supplementary Materials provide, respectively, a detailed breakdown of these adjustments by specialty and the number of cases written per author.

\subsubsection{Specific Use Cases}

In addition to the documents from the initial distribution, four specific sets of reports were assembled. Each set was designed to address a specific clinical information extraction use case:

\begin{description}
    \item[Coding] Surgery reports from digestive surgery, orthopedic surgery, traumatology, and urology were specifically collected for a use case on ICD-10 diagnostic coding.
    \item[Identifying biomarkers in oncology] Pathology reports containing descriptions of tumor biomarkers used to inform diagnosis, prognosis, and targeted therapy selection in oncology: tissue and genomic alterations such as mutations, amplifications, and protein expression. The use case for this dataset is the automatic identification of these biomarkers.
    \item[Identifying the response to treatment in oncology] Oncology consultation reports mentioning treatment response (complete response, partial response, stable disease, progressive disease, not applicable, or indeterminate). The associated use case aims at classifying RECIST-style information from these reports.
    \item[Infectiology] Reports describing infectious episodes (including bacteremia) along with the causative bacteria and the primary site of infection. 
\end{description}

Other use cases (pseudonymization and summarization) are planned for the PARTAGES project (described below). However, the documents dedicated to these tasks do not require a distribution that differs from that described in the previous section.%

\subsection{Implementation}


The PARTAGES project, funded by the French government under the France 2030 initiative (operated by Bpifrance), brings together a consortium of 32 partners, including research teams, public and private healthcare institutions, and AI-focused deeptech companies. Its aim is to develop open resources to support the emergence of generative AI solutions in healthcare. The creation of the PARHAF corpus of clinical reports is one of the consortium's initiatives.

The corpus development was initiated through a scoping phase involving NLP experts and physicians, aimed at balancing production volume with budgetary constraints. Writing time was estimated at 60 minutes per document: 45 minutes for the first author and 15 minutes for review, validation, and correction by a second expert. This estimation was based on the expertise of the physicians involved and consultation with twelve residents from different specialties, active within their respective residents’ associations. A maximum duration of 60 minutes per document was retained, with a gross hourly compensation of~\euro 40, corresponding to a maximum of \euro 40 per completed and reviewed report. Recruitment was conducted through a temporary employment agency.

A national outreach campaign targeted 21 residents’ associations across different specialties in France. Eleven associations disseminated the call for participation, representing the following specialties: internal medicine, infectious diseases, visceral surgery, obstetrics and gynecology, neurology, pulmonology, public health, urology, oncology, anesthesiology and intensive care, and anatomical pathology. The call for participation was further circulated via the project’s hospital partners and their associated medical networks, enabling the inclusion of residents from additional specialties: nephrology, hematology, orthopedics, pediatrics, gastroenterology and hepatology, and cardiology. More than 500 applications were received, reflecting strong engagement from the medical community. A final panel of 104 residents was selected, prioritizing residents in the later stages of training and ensuring broad geographic representation. This response confirmed residents' commitment to developing digital commons and supporting generative AI projects in healthcare.

To ensure consistency and quality of the outputs, a structured support framework was implemented, including regular webinars, methodological guidelines, a centralized communication platform, and dedicated support. Contributors were also involved in methodological refinements through specialty-specific meetings, enabling adaptation of instructions to clinical practice and ensuring the validity and representativeness of the corpus.

From a financial perspective, operational costs primarily consisted of physician compensation, increased by the management fees of the temporary employment agency (multiplicative coefficient of 1.9 applied to the gross remuneration). The production of the 7,394 clinical reports, totaling 5,518 hours of effective work, resulted in a total operational cost of approximately \euro 495,000.

\section{Data Records}



\corpusabbr\ consists of a single JSON file containing structured metadata about fictitious patients and the clinical documents associated with them. Each entry in the data array corresponds to one patient record and includes patient-level metadata, contextual information about the care scenario, and a list of associated documents.%

The documents themselves are not embedded in the JSON file. Instead, each document is referenced via a relative file path pointing to an external text file. These text files are stored separately and organized by medical specialty, with one directory per specialty. Each document file contains raw, unannotated plain text in French, with no markup, labels, or structural tags applied.

The JSON file, therefore, acts as the index and metadata layer of the corpus, while the directory structure contains the raw textual content. The linkage between metadata and text relies exclusively on the relative paths specified in the \texttt{documents[].path} fields.

The high-level structure of the JSON format and the path-based schema description are described in Figure~\ref{tab:structure} and Table~\ref{tab:description}, respectively.

In addition to this standalone dataset, we also distribute a Hugging Face dataset (Parquet/Arrow) that is a derived representation generated automatically from the JSON files. Both formats, therefore, contain identical information and differ only in storage layout.

The PARHAF corpus is openly available under the Creative Commons Attribution~4.0 International (CC BY 4.0) license and the Etalab 2.0 license. It was released on March 25, 2026. The primary distribution is on Hugging Face (\url{https://huggingface.co/datasets/HealthDataHub/PARHAF}). 

\begin{figure}
\footnotesize
\dirtree{%
.1 root (object).
.2 name: string.
.2 version: string.
.2 licenses: array of string.
.2 license\_urls: array of string.
.2 description: string.
.2 patient\_count: integer.
.2 document\_count: integer.
.2 data: array of object.
.3 \textit{element (patient)}.
.4 id: string.
.4 local\_id: string.
.4 specialty: string.
.4 author: string.
.4 reviewer: string.
.4 pool: string.
.4 suggested\_scenario: object.
.5 name: string.
.5 age: object.
.6 value: integer.
.6 unit: string.
.5 sex: string.
.5 admission\_mode: string (optional).
.5 discharge\_mode: string (optional).
.5 primary\_procedure: object (optional).
.6 code: string.
.6 description: string.
.5 primary\_diagnosis: object.
.6 code: string.
.6 description: string.
.5 type\_of\_care: string (optional).
.4 documents: array of object.
.5 \textit{element (document)}.
.6 type: string.
.6 header: string.
.6 path: string.
.6 word\_count: integer.
.4 structured\_abstract: object (optional).
.5 primary\_diagnosis: array of object (optional).
.6 \textit{element (diagnosis)}.
.7 code: string (optional).
.7 description: string.
.5 primary\_procedure: array of object.
.6 \textit{element (procedure)}.
.7 code: string.
.7 description: string.
.5 admission\_mode: string (optional).
.5 discharge\_mode: string (optional).
.5 length\_of\_stay: object (optional).
.6 value: integer.
.6 unit: string.
}
\caption{High-level structure of the dataset (JSON format)}
\label{tab:structure}
\end{figure}

\begin{table}
\centering
\small
\begin{tabular}{>{\ttfamily}llp{7cm}}
\toprule
\textnormal{\textbf{Path}} & \textbf{Type} & \textbf{Description} \\
\midrule
\multicolumn{3}{l}{\textbf{\textit{Root level}}} \\
\midrule
name & string & Name of the corpus \\
version & string & Version \\
licenses & string & Licenses of the dataset \\
license\_urls & string & URL to license descriptions \\
description & string & Short description of the corpus \\
patient\_count & integer & Number of patients in the corpus \\
document\_count & integer & Number of documents (reports) in the corpus \\
data[] & array  & List of patients \\
\midrule
\multicolumn{3}{l}{\textbf{\textit{data[] (one patient)}}} \\
\midrule
id & string & Global unique patient identifier  \\
local\_id & string & Local id within the specialty  \\
specialty & string & Medical specialty  \\
author & string & Author trigram  \\
reviewer & string & Reviewer trigram  \\
pool & string & Dataset partition. Some patients are part of specific use cases (UC1--6). \\
suggested\_scenario & dict & Basic scenario suggested to the report author \\
documents[] & array  & List of reports for this patient\\
structured\_abstract & dict & Optional and uncurated abstract filled by the author. May not reflect the real content of the report.  \\
\midrule

\multicolumn{3}{l}{\textbf{\textit{suggested\_scenario (object)}}}\\

\midrule
name & string & Patient name (fictional)  \\
age.value & integer & Age value  \\
age.unit & string & Age unit \\
sex & string & Patient sex  \\
admission\_mode & string & Admission source  \\
discharge\_mode & string & Discharge destination  \\
primary\_procedure.code & string & CCAM code\\
primary\_procedure.description & string & \PZi{Procedure} label   \\
primary\_diagnosis.code & string & ICD10 code  \\
primary\_diagnosis.description & string & Diagnosis label  \\
type\_of\_care & string & Care description  \\
\midrule

\multicolumn{3}{l}{\textbf{\textit{documents[]  (between 1 and 3 patient reports})}} \\
\midrule
type & string & Document type \\
header & string & Document title  \\
word\_count & integer & Number of words in the report \\
path & string & Relative path to raw text  \\
\bottomrule
\end{tabular}

\caption{Path-based schema description}
\label{tab:description}
\end{table}

\section{Data Overview}



\label{sec:data_overview}

A subset of the dataset is temporarily embargoed to enable future evaluations under controlled conditions, thereby limiting the risk of large language model contamination through prior exposure to the data. In total, we release \patientcounttrain~patients (\doccounttrain~documents) and keep \patientcounttest~patients (\doccounttest~documents) under embargo for future release.

\begin{figure}
    \centering \includegraphics[width=\linewidth]{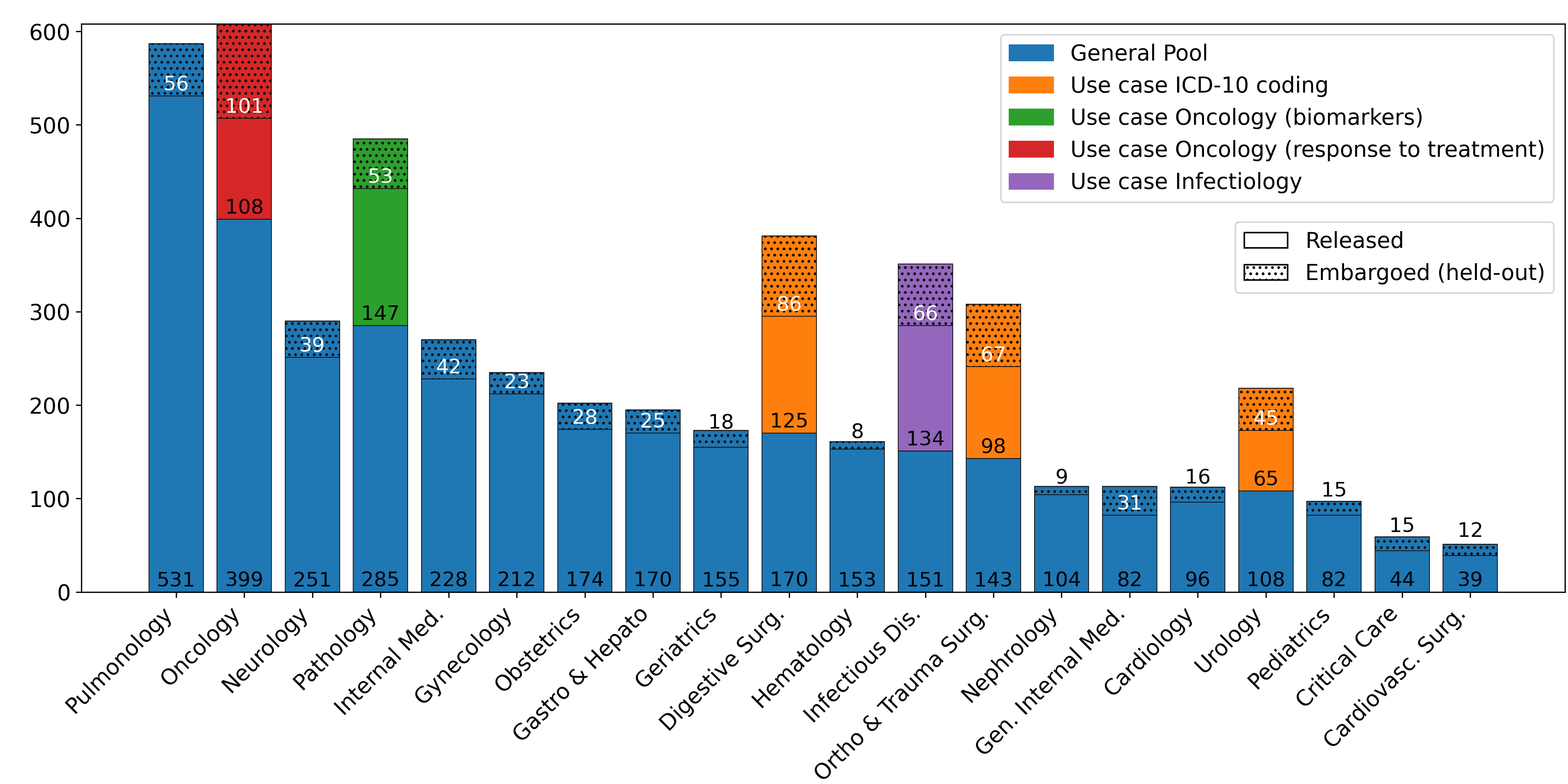}
\caption{Patient count per specialty. Specific documents, off distribution, were written for four use cases. A subset of the dataset is temporarily embargoed to enable future evaluations under controlled conditions, thereby limiting the risk of large language model contamination through prior exposure to the data. In total, we release \patientcounttrain~patients (\doccounttrain~documents) and keep \patientcounttest~patients (\doccounttest~documents) under embargo for future release.}
    \label{fig:patient_distrib}
\end{figure}

\begin{figure}
    \centering
    \includegraphics[width=0.7\linewidth]{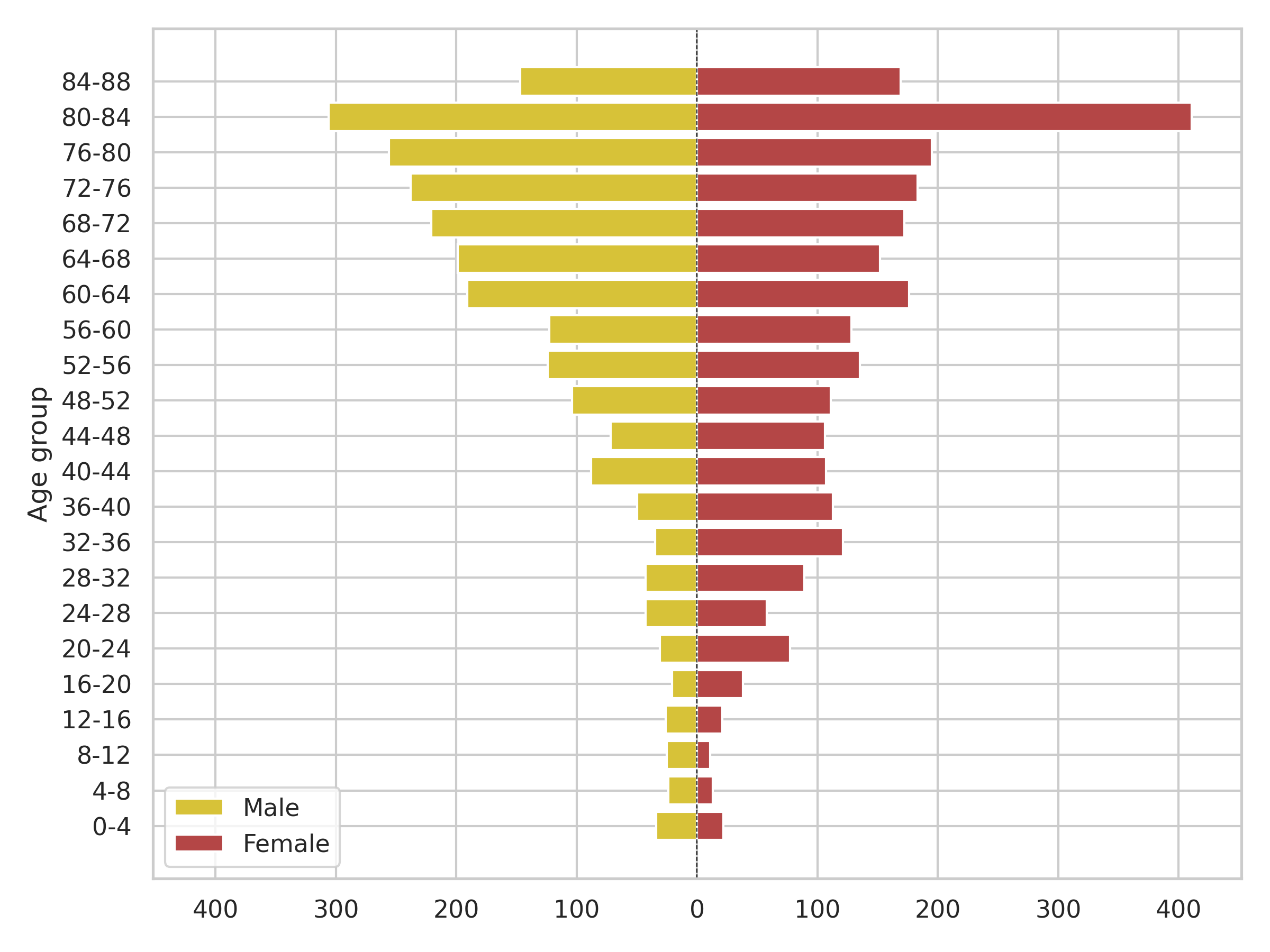}
    \caption{\corpusabbr\ population pyramid chart.}
    \label{fig:population_pyramid}
\end{figure}

\begin{figure}
    \centering
    \includegraphics[width=0.6\linewidth]{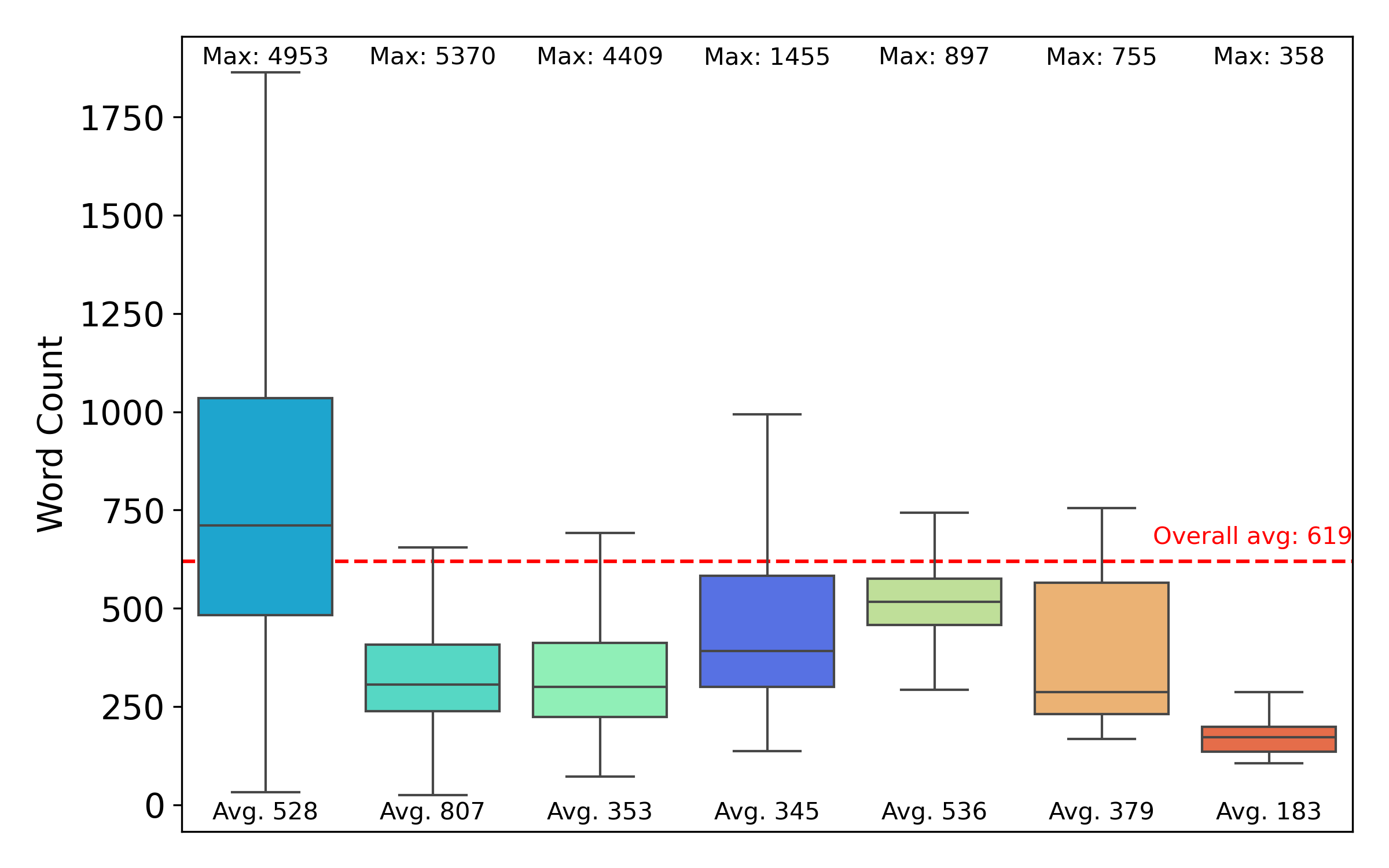}
    \includegraphics[width=0.6\linewidth]{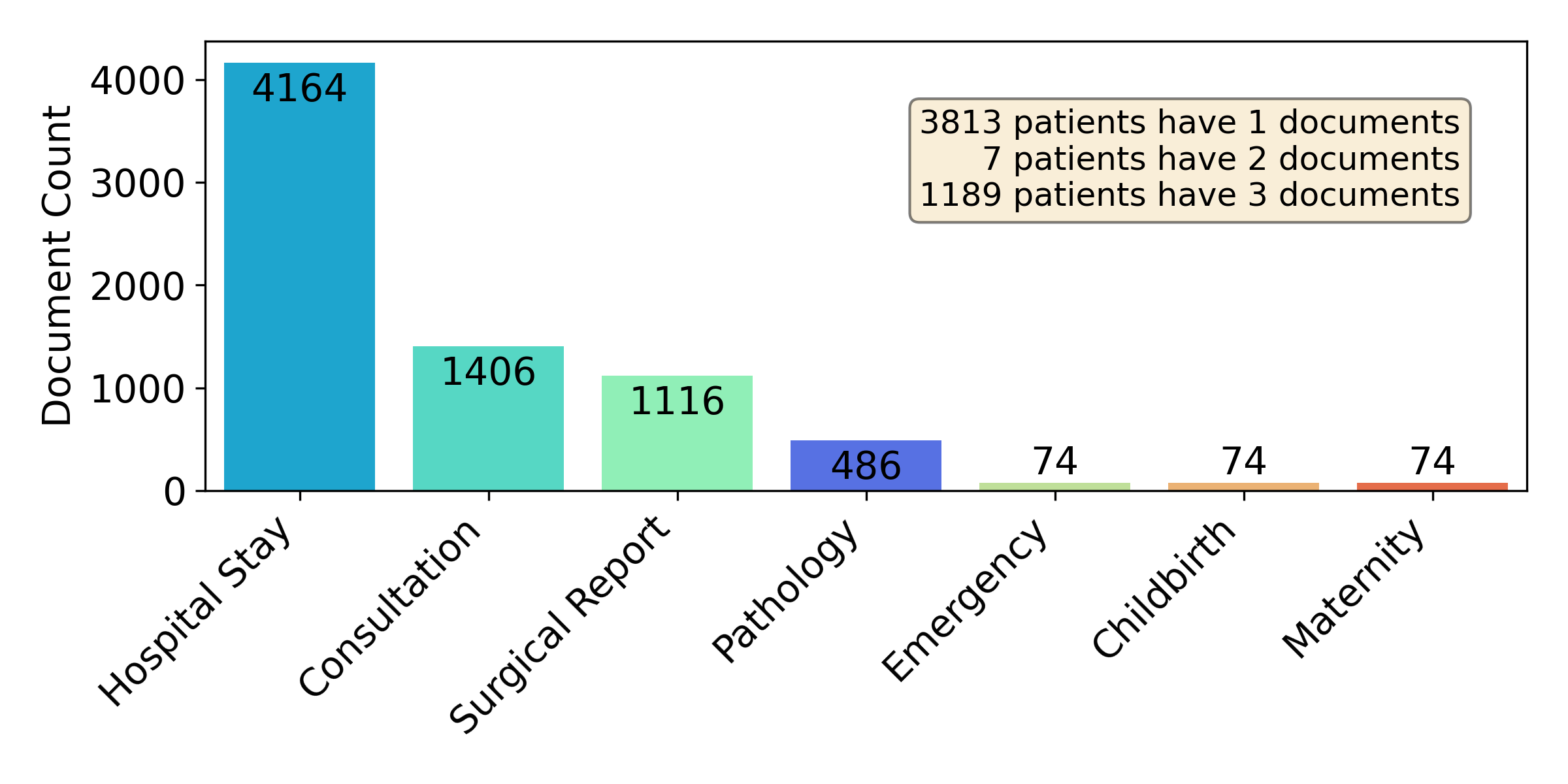}
    \caption{Document and word count by document type. The number of documents to write for a patient was imposed by the initial scenario. Word count is computed by the library \textit{edsnlp} (v0.20).}        
    \label{fig:document_types}
\end{figure}

Figures~\ref{fig:patient_distrib}, \ref{fig:population_pyramid}, and~\ref{fig:document_types} provide details about, respectively, the patient count per medical specialty, the population pyramid chart, and the document and word counts by document type.

\section{Technical Validation}


All contributors involved in report writing and validation completed standardized online training covering the study objectives, authoring guidelines, and validation procedures. The full clinical scenario framework was presented in detail to ensure a consistent understanding of context and expectations. Additional specialty-specific instructions were provided where relevant, including requirements regarding the number and types of documents per patient and scenario-dependent constraints.

To preserve ecological validity and avoid burdening contributors with rigid formatting that would diverge from real-world clinical practice, reports were produced as guided but free-text documents within an online text-editing environment. Human validators were responsible for content quality control, assessing clinical coherence and adherence to the instructions. This step resulted in a rejection rate of 3.6\% of submitted documents.

Beyond human review, an automated ``sanity check'' pipeline was implemented to verify compliance with key structural and procedural requirements. These checks included validating document type and the expected number of documents per patient, confirming final hospitalization duration entries, and recording required clinical procedures or acts. This automated stage ensured the consistency of core structured elements across the corpus.

Following automated validation, 120 documents required manual correction. These interventions addressed minor typographical errors within structured fields, format-style deviations that interfered with automated parsing, or unauthorized modifications to the original instruction template. No corrections altered the underlying clinical content; all changes were limited to restoring technical conformity with the dataset specifications.



\section{Data Availability}






The PARHAF corpus is openly available under the Creative Commons Attribution~4.0 International (CC BY 4.0) license, and the Etalab 2.0 license. It was released on March 25, 2026. The primary distribution is on Hugging Face at \url{https://huggingface.co/datasets/HealthDataHub/PARHAF}. 

\section{Code Availability}

The code used for data processing and quality control is publicly available on GitHub at \url{https://github.com/xtannier/PAHRAF_cleaning_and_publication}. This includes scripts for converting source documents from .docx to plain text, generating JSON metadata, building the Hugging Face Parquet dataset, and running the automated sanity-check pipeline described above.


\bibliographystyle{naturemag}
\bibliography{biblio}

\section{Author Contributions}


\textbf{Xavier Tannier}: Conceptualization, Methodology, Software, Validation, Data Curation, Writing - Original Draft, Writing - Review \& Editing, Visualization, Funding acquisition.
\textbf{Salam Abbara}: Conceptualization, Methodology, Writing - Review \& Editing, Funding acquisition.
\textbf{Rémi Flicoteaux}: Conceptualization, Methodology, Validation, Writing - Review \& Editing.
\textbf{Youness Khalil}: Methodology, Writing - Review \& Editing, Project administration.
\textbf{Aurélie Névéol}: Conceptualization, Writing - Review \& Editing, Funding acquisition.
\textbf{Pierre Zweigenbaum}: Conceptualization, Writing - Review \& Editing, Funding acquisition.
\textbf{Emmanuel Bacry}: Conceptualization, Supervision, Project coordinator, Funding acquisition, Writing - Review \& Editing.

\section{Competing Interests}

The authors declare no competing interests related to this work.

\section{Acknowledgements}
We thank the authors of the reports for their contribution and feedback on the protocol, as well as the PARTAGES consortium members for fruitful discussions towards corpus development. We also thank Florian Pons for helping with operational support and project coordination.

\section{Funding}

This work was carried out as part of the PARTAGES project, awardee of the Bpifrance France~2030 call for proposals ``Digital Commons for Generative Artificial Intelligence.''

\section{Ethics statement*}
Through their affiliations with French public service agencies, the developers of the PARHAF corpus have benefited from access to SNDS data. The clinical scenarios used to write the clinical documents in the PARHAF corpus are based on aggregated public health statistics and do not pertain to identifiable real patients. No private information about individual subjects was used in this study; therefore, no IRB or ethics approval was required to create or distribute the PARHAF corpus.

The clinical document authors involved in this study were 
apprised of the full document creation protocol. Participation was voluntary, and authors were compensated for their work in accordance with French labor laws.

The PARHAF corpus is intended for use as educational material and as support for the development and evaluation of clinical NLP systems. It is not intended for clinical use. 

\appendix
\setcounter{figure}{0}
\renewcommand\thefigure{S\arabic{figure}}    

\section*{Supplementary Material}

\begin{figure}[h]
    \centering
    \includegraphics[width=0.9\linewidth]{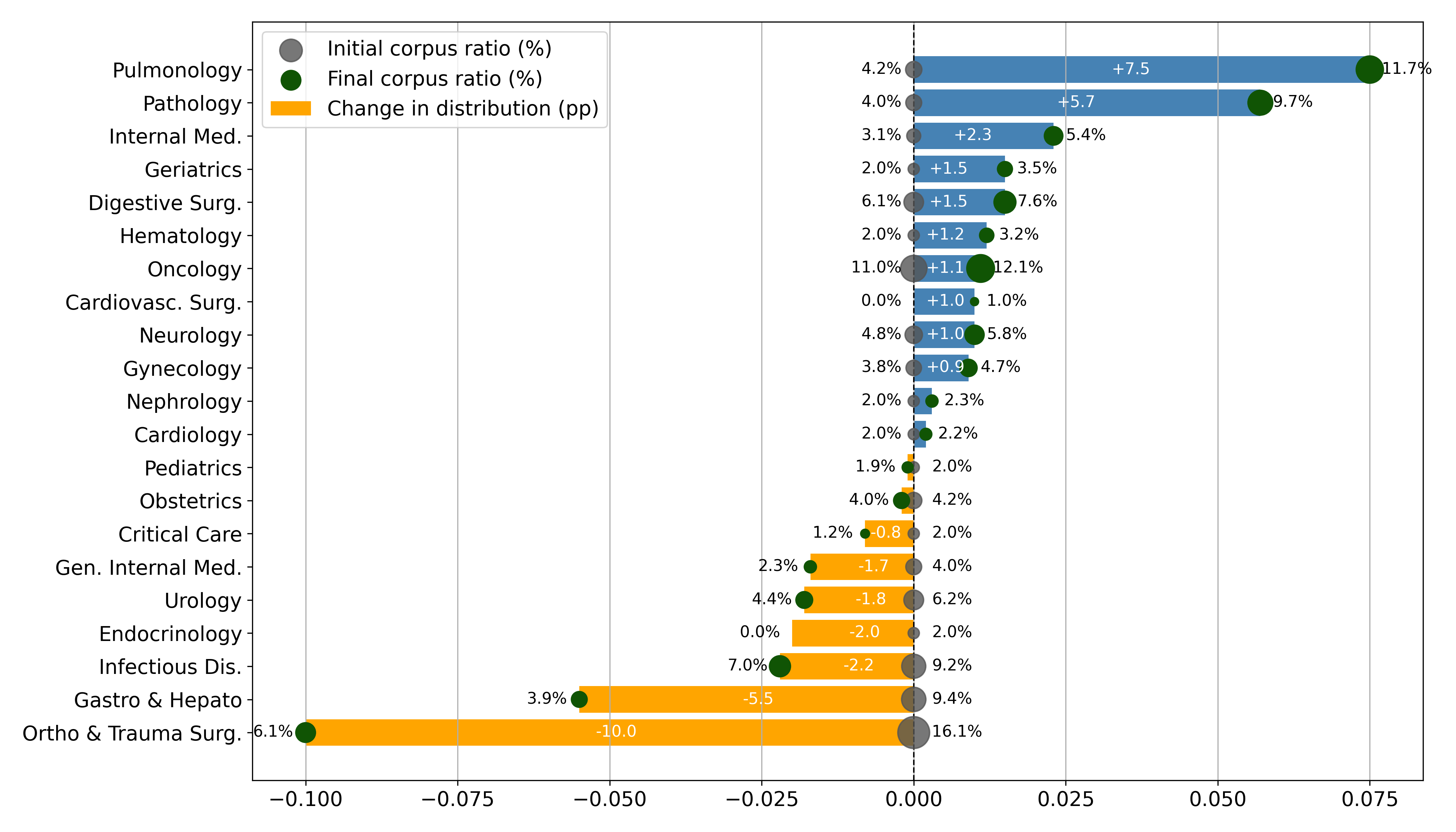}
    \caption{Change in distribution for each specialty, after the adjustment described in Section~\ref{sec:core_distribution}.}
    \label{fig:change_in_distribution}
\end{figure}

\begin{figure}[h]
    \centering
    \includegraphics[width=0.23\linewidth]{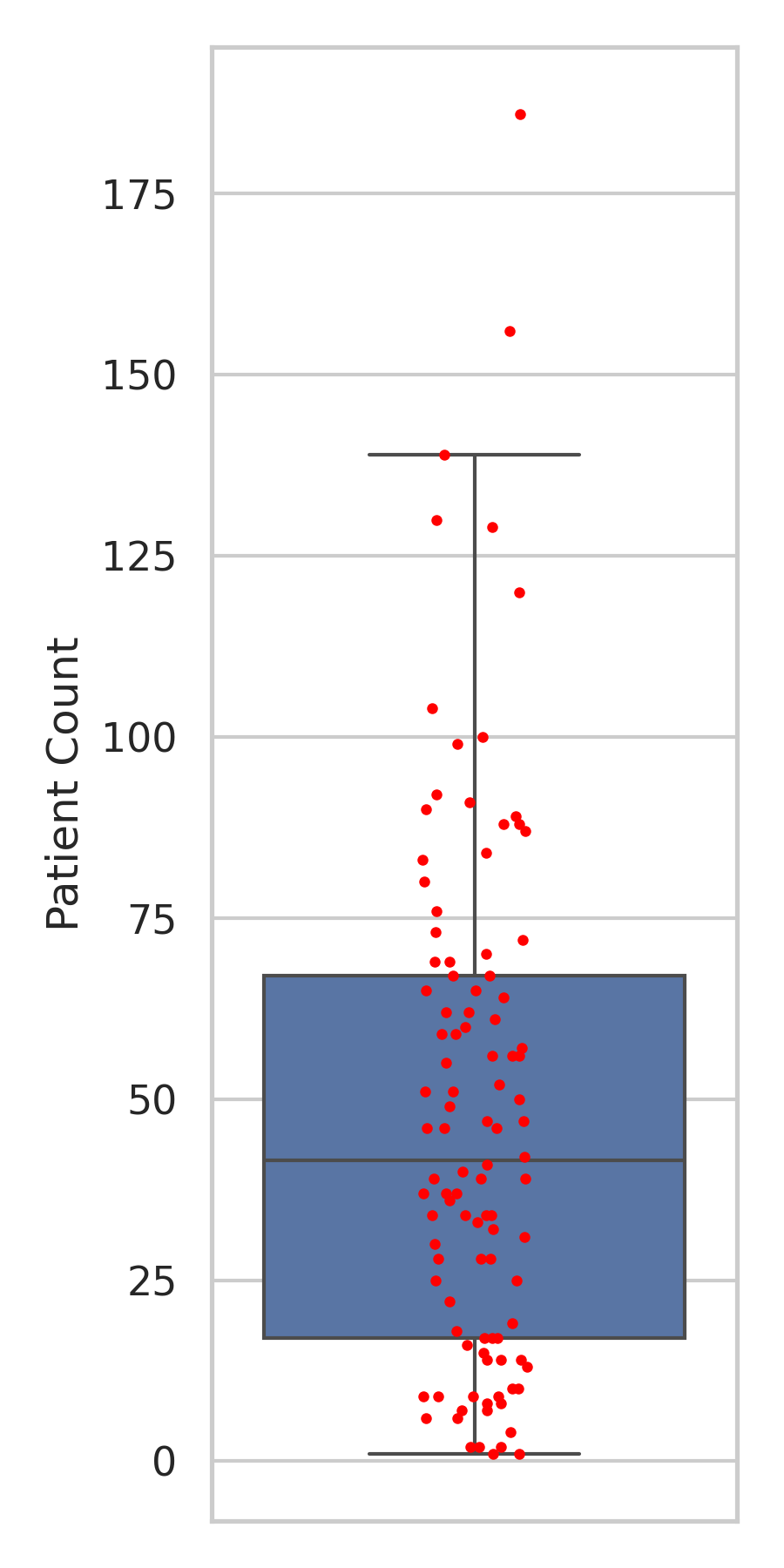}
    \caption{Patient counts by author}
    \label{fig:count_per_author}
\end{figure}

Figure~\ref{fig:change_in_distribution} shows the changes in distribution for each specialty, after the adjustment described in Section~\ref{sec:core_distribution}. Figure~\ref{fig:count_per_author} illustrates the number of patients written by each author.

\end{document}